# Enabling and Inhibitory Pathways of University Students' Willingness to Disclose AI Use: A Cognition–Affect–Conation Perspective

**First Author and Corresponding Author**
Yiran Du
University of Cambridge, Cambridge, UK
yd392@cam.ac.uk

**Second Author**
Huimin He
Xi'an Jiaotong-Liverpool University, Suzhou, China
Huimin.he@xjtlu.edu.cn

**Abstract**
The increasing integration of artificial intelligence (AI) in higher education has raised important questions regarding students' transparency in reporting AI-assisted work. This study investigates the psychological mechanisms underlying university students' willingness to disclose AI use by applying the Cognition–Affect–Conation (CAC) framework. A sequential explanatory mixed-methods design was employed. In the quantitative phase, survey data were collected from 546 university students and analysed using structural equation modelling to examine the relationships among cognitive perceptions, affective responses, and disclosure intention. In the qualitative phase, semi-structured interviews with 22 students were conducted to further interpret the quantitative findings. The results indicate that psychological safety significantly increases students' willingness to disclose AI use and is positively shaped by perceived fairness, perceived teacher support, and perceived organisational support. Conversely, evaluation apprehension reduces disclosure intention and psychological safety, and is strengthened by perceived stigma, perceived uncertainty, and privacy concern. Qualitative findings further reveal that institutional clarity and supportive instructional practices encourage openness, whereas policy ambiguity and fear of negative evaluation often lead students to adopt cautious or strategic disclosure practices. Overall, the study highlights the dual role of enabling and inhibitory psychological mechanisms in shaping AI-use disclosure and underscores the importance of supportive institutional environments and clear guidance for promoting responsible AI transparency in higher education.

## 1. Introduction

Artificial intelligence (AI) technologies are increasingly integrated into higher education, transforming how students access information, generate content, and complete academic tasks. Tools based on generative AI and large language models enable students to obtain explanations, draft written material, and support problem-solving activities, thereby becoming embedded in everyday learning practices (Kasneci et al., 2023; Lee et al., 2026). While these technologies offer opportunities to enhance learning efficiency and access to knowledge, they also raise concerns related to academic integrity, transparency, and responsible use (Doğan et al., 2025; Jensen et al., 2025).

One emerging issue in this context is students' willingness to disclose their use of AI tools when completing academic work. Disclosure, such as acknowledging AI assistance in assignments or reporting AI-supported work to instructors, is increasingly viewed as an important aspect of ethical AI use in education (Ans et al., 2026; Wiese et al., 2025). However, students may hesitate to disclose AI use due to concerns about negative evaluation, policy ambiguity, or social judgement. Existing research has largely focused on AI adoption or detection of AI-generated content, with less attention given to the psychological factors that influence students' decisions to disclose AI use (Yan et al., 2025).

To address this gap, this study examines the enabling and inhibitory mechanisms underlying university students' willingness to disclose AI use through the Cognition–Affect–Conation (CAC) framework.

This perspective conceptualises behaviour as a process in which cognitive evaluations shape emotional responses that subsequently influence behavioural intentions (Zhou & Zhang, 2024). Drawing on this framework, the study investigates how institutional perceptions, including fairness, teacher support, organisational support, stigma, uncertainty, and privacy concerns, influence affective states such as psychological safety and evaluation apprehension, which ultimately shape students' willingness to disclose AI use.

## 2. Literature Review
### 2.1 AI in Higher Education
Artificial intelligence (AI) technologies have rapidly expanded within higher education, reshaping teaching, learning, and assessment practices (Qian, 2025). Tools based on natural language processing, machine learning, and generative models increasingly support activities such as automated feedback, personalised learning pathways, and content generation (Lee et al., 2026). The emergence of large language model–based systems such as ChatGPT has particularly accelerated the integration of AI into academic contexts, allowing students to obtain explanations, draft texts, and assist with problem-solving tasks (Li et al., 2025). Consequently, AI is becoming embedded in routine academic workflows across disciplines (Kasneci et al., 2023).

Scholarly research has examined both the pedagogical opportunities and the challenges associated with AI adoption in higher education. On one hand, AI tools can enhance learning efficiency, provide adaptive support, and facilitate access to information (Jensen et al., 2025). On the other hand, concerns have emerged regarding academic integrity, overreliance on automated assistance, and the transparency of AI-mediated work (Doğan et al., 2025). Institutions and instructors are therefore increasingly exploring policies that regulate how AI tools may be used and how such use should be reported in academic assignments (Francis et al., 2025).

Within this evolving context, student behaviour regarding AI use has become an important research focus (Nzenwata et al., 2024). While many students experiment with AI tools to support learning tasks, their perceptions of legitimacy, risk, and institutional expectations vary considerably (Kasneci et al., 2023). These perceptions could shape how students integrate AI into academic work and whether they openly acknowledge such use (Çerkini et al., 2025). Understanding these dynamics requires examining not only technological adoption but also the psychological processes underlying students' decisions about disclosure.

### 2.2 Willingness to Disclose AI Use
Willingness to disclose AI use refers to the extent to which students voluntarily report or acknowledge their use of AI tools when completing academic tasks (Garcia Ramos, 2025). Disclosure may occur in different forms, such as citing AI assistance in assignments, informing instructors about AI-generated contributions, or following institutional policies that require transparency in AI-assisted work (Ans et al., 2026). In the context of emerging AI technologies, disclosure behaviour is increasingly viewed as a key indicator of responsible and ethical AI use in education (Wiese et al., 2025).

Existing literature suggests that disclosure behaviour is influenced by multiple psychological and contextual factors (Zhou & Wu, 2025b). Students may weigh perceived benefits of transparency against potential risks, such as academic penalties or negative instructor evaluations (Stone, 2025). Cognitive evaluations (e.g., perceptions of policy clarity or fairness), affective responses (e.g., anxiety or trust), and social norms within academic environments can all shape students' intentions to disclose AI use (Zhou & Wu, 2025a). These factors reflect broader decision-making processes associated with ethical reporting and self-regulation in academic contexts (Ans et al., 2026).

Despite growing interest in AI use in education, empirical research specifically addressing students' willingness to disclose AI assistance remains limited. Much of the current literature focuses on academic integrity violations or detection mechanisms rather than the psychological processes underlying voluntary disclosure (Yan et al., 2025). As AI tools become more prevalent, examining the enabling and inhibitory pathways that influence disclosure decisions, particularly from integrated psychological perspectives such as cognition, affect, and conation, becomes essential for developing effective policies and educational practices (Đerić et al., 2025).

## 3. Theoretical Framework

This study draws on the Cognition–Affect–Conation (CAC) framework to explain university students' willingness to disclose their use of AI in academic contexts. The CAC framework conceptualises behaviour as a sequential psychological process in which individuals' cognitive evaluations influence their affective responses, which subsequently shape behavioural intentions (Zhou & Wang, 2025). This framework has been widely used in behavioural and educational research because it captures how individuals interpret situational cues, experience emotional reactions, and ultimately decide whether to engage in a particular behaviour (Zhou & Zhang, 2024). In the context of AI use in higher education, disclosure decisions are unlikely to be purely rational or purely emotional; instead, they result from the interaction between students' perceptions of their academic environment and their emotional responses to potential evaluation or judgement. The CAC framework therefore provides an appropriate theoretical lens for understanding how students form intentions to disclose AI use.

Applying the CAC framework allows this study to examine both enabling and inhibiting psychological pathways that influence disclosure behaviour. Specifically, cognitive perceptions related to the academic environment may generate affective responses that either facilitate or discourage openness about AI use. Positive perceptions may create emotional conditions that support transparency, whereas negative perceptions may evoke concerns about potential consequences or social judgement, thereby discouraging disclosure. Based on this perspective, the conceptual model proposes that cognitive factors, perceived fairness, perceived teacher support, perceived organisational support, perceived stigma, perceived uncertainty, and privacy concern, shape affective states, namely psychological safety and evaluation apprehension, which in turn influence students' willingness to disclose AI use. The conceptual model illustrating these relationships is presented in Figure 1.

**Figure 1. The Conceptual Model**

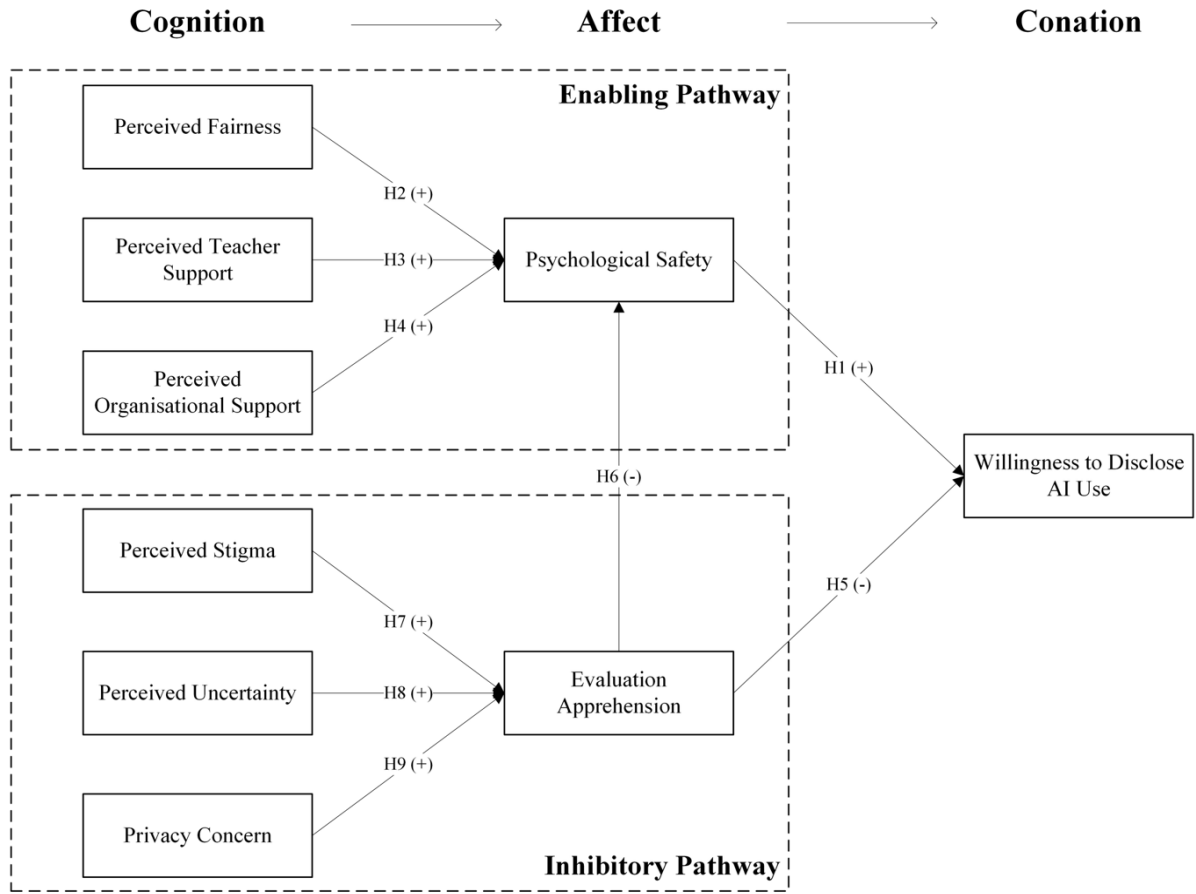

## 4. Research Questions and Hypothesis Development
### 4.1 Enabling Pathway of Willingness to Disclose AI Use
Based on the conceptual model, the first research question (RQ1) asks how psychological safety, perceived fairness, perceived teacher support, and perceived organisational support influence university students' willingness to disclose AI use. Psychological safety refers to the perception that individuals can express their behaviours, ideas, or concerns without fear of negative consequences such as punishment, embarrassment, or unfair judgement (Edmondson & Bransby, 2023). Prior research in organisational and educational contexts shows that psychological safety facilitates open communication, information sharing, and behavioural transparency (Vella et al., 2024). In learning environments, students who experience higher psychological safety are more likely to engage in discussions, ask questions, and acknowledge their learning practices (Somerville et al., 2023). In the context of AI-assisted learning, students who feel psychologically safe may be more willing to disclose their use of AI tools when completing academic tasks (McGuire, 2025).

Perceived fairness is an important cognitive factor that may influence students' psychological safety. Perceived fairness refers to students' evaluations of whether institutional policies and assessment practices are equitable and transparent (Lünich et al., 2024). Research on organisational justice suggests that fairness perceptions significantly influence individuals' emotional responses and trust in institutions (Malhotra et al., 2022). When individuals believe that procedures and decisions are fair, they are more likely to feel secure and confident when interacting with authorities (Chambers et al., 2023). In higher education, if students perceive policies regarding AI use and academic evaluation as fair, they may feel less concerned about being unfairly judged for using AI tools, thereby enhancing their psychological safety (Stone, 2025).

Perceived teacher support may also contribute to the development of psychological safety. Perceived teacher support reflects the extent to which students believe that instructors provide guidance, encouragement, and understanding in the learning process (Tao et al., 2022). Empirical studies in educational psychology indicate that supportive instructor–student relationships can enhance students' trust, engagement, and willingness to communicate about their academic experiences (Held & Mori, 2024). When instructors demonstrate openness and constructive attitudes toward the use of emerging technologies, students may feel more comfortable discussing their learning practices, including their use of AI tools (Prananto et al., 2025).

Perceived organisational support refers to students' beliefs that the university values their learning needs and provides adequate guidance and resources (Orpina et al., 2022). In organisational behaviour research, perceived organisational support has been shown to strengthen individuals' trust in institutions and foster positive emotional experiences (Anton, 2025). In the context of higher education, universities that provide clear policies, guidance, and support regarding AI use may reduce ambiguity and create a more supportive learning environment (Jeilani & Abubakar, 2025). Such institutional support may enhance students' confidence that their actions will be understood and treated fairly, thereby strengthening their psychological safety (Dalban et al., 2025). Accordingly, the following hypotheses are proposed:

H1: Psychological safety is positively associated with willingness to disclose AI use.
H2: Perceived fairness is positively associated with psychological safety.
H3: Perceived teacher support is positively associated with psychological safety.
H4: Perceived organisational support is positively associated with psychological safety.

**4.2 Inhibitory Pathway of Willingness to Disclose AI Use**
Based on the conceptual model, the second research question (RQ2) examines how evaluation apprehension, perceived stigma, perceived uncertainty, and privacy concern influence university students' willingness to disclose AI use. Evaluation apprehension refers to individuals' anxiety about being negatively judged when their behaviour is evaluated by others (Ismail & Heydarnejad, 2023). Research suggests that evaluation apprehension can discourage individuals from expressing ideas, admitting mistakes, or revealing behaviours that might be socially sensitive (McGaghie, 2018). In academic contexts, students who worry about negative judgement from instructors or peers may avoid disclosing behaviours that could affect their academic reputation (Prasad et al., 2023). In the case of AI-assisted learning, such apprehension may reduce students' willingness to openly report their use of AI tools. In addition, prior studies indicate that evaluation apprehension can undermine individuals' sense of psychological safety because concerns about judgement or criticism reduce feelings of interpersonal security and openness (Geerts et al., 2021). When students anticipate negative evaluation, they may feel less safe communicating their learning behaviours, thereby weakening psychological safety.

Perceived stigma may contribute to the development of evaluation apprehension. Stigma refers to the perception that a particular behaviour is socially disapproved of or associated with negative moral judgement (Troup et al., 2022). In higher education, AI use in academic work may sometimes be viewed as inappropriate or academically questionable, especially when institutional policies remain unclear (Yan et al., 2025). Research on stigma in social and educational contexts shows that when individuals believe a behaviour is stigmatised, they become more sensitive to others' evaluations and may experience greater anxiety about how they will be perceived (Orta-Aleman et al., 2026). Consequently, if students perceive that using AI tools carries social stigma within their academic environment, they may experience stronger evaluation apprehension when considering whether to disclose such use.

Perceived uncertainty may also increase evaluation apprehension. Uncertainty arises when individuals lack clear information about rules, expectations, or possible consequences of their actions (Usher & Barak, 2024). In many universities, policies regarding AI use are still evolving, and students may not fully understand when AI use is acceptable or how it should be reported (Garcia Ramos, 2025). Prior research on uncertainty in organisational and educational settings suggests that ambiguous policies can increase anxiety and cautious behaviour because individuals fear making mistakes or violating expectations (Zhu et al., 2025). Under such conditions, students may become more concerned about how their behaviour will be evaluated, thereby increasing evaluation apprehension (Stone, 2025).

Privacy concern represents another factor that may influence evaluation apprehension. Privacy concern refers to individuals' worries about the exposure or misuse of personal information (Herriger et al., 2025). Research on technology use indicates that privacy concerns can generate anxiety and defensive behaviour when individuals interact with digital systems or disclose personal information (Zheng et al., 2025). In the context of AI use disclosure, students may worry that reporting their AI use could expose their learning processes to closer scrutiny or monitoring by instructors and institutions (Hu & Min, 2023). Such concerns may heighten sensitivity to evaluation and contribute to stronger evaluation apprehension (Carmody et al., 2021). Accordingly, the following hypotheses are proposed:

H5: Evaluation apprehension is negatively associated with willingness to disclose AI use.
H6: Evaluation apprehension is negatively associated with psychological safety.
H7: Perceived stigma is positively associated with evaluation apprehension.
H8: Perceived uncertainty is positively associated with evaluation apprehension.
H9: Privacy concern is positively associated with evaluation apprehension.

## 5. Methods
### 5.1 Research Design
This study adopted a sequential explanatory mixed-methods design, in which quantitative data collection and analysis were followed by qualitative inquiry to further interpret the quantitative findings (Cohen et al., 2018). In the first phase, a questionnaire survey was administered to examine the relationships among the constructs in the proposed conceptual model of students' willingness to disclose AI use, focusing on the interplay between cognitive factors, affective responses, and behavioural intention. The quantitative results provided an overall picture of students' perceptions and the hypothesised relationships among these variables. In the second phase, semi-structured interviews were conducted with a purposively selected subset of survey participants to gain deeper insights into students' experiences and perspectives regarding AI use and disclosure in academic contexts. The qualitative findings were used to elaborate and contextualise the quantitative results, thereby providing a more comprehensive understanding of the factors influencing students' willingness to disclose AI use.

### 5.2 Participants and Sampling
A total of 572 students initially responded to the questionnaire. After data screening, 26 responses were excluded due to careless responding (Ward & Meade, 2023), resulting in a final valid sample of 546 participants for analysis. Participants were recruited from multiple universities across different regions of China using a multi-site convenience sampling strategy. Survey invitations were disseminated through course instructors, university mailing lists, and student social media groups. Interested students accessed the survey via an online link and completed the questionnaire voluntarily. The demographic characteristics of the questionnaire participants are summarised in Appendix A. The sample comprised 294 female students (53.8%) and 252 male students (46.2%). In terms of age distribution, 36.3% of participants were aged 18–20, 47.1% were aged 21–23, and 16.7% were aged 24 or above. Regarding study level, 77.1% of respondents were undergraduate students and 22.9% were postgraduate students. Participants represented diverse disciplinary backgrounds, with 50.2% enrolled in STEM programmes and 49.8% in non-STEM fields.

For the qualitative phase, interview participants were drawn from questionnaire respondents who indicated their willingness to participate in a follow-up interview. A purposive sampling strategy with maximum variation was employed to capture diverse perspectives on students' willingness to disclose AI use. Selection prioritised variation in disclosure intention and subsequently considered diversity in gender, age, study level, and disciplinary background. Disclosure intention was determined using participants' scores on the disclosure intention scale included in the questionnaire. For each respondent, a mean score across the scale items was calculated, and participants whose scores fell within the top quartile of the distribution were classified as having high willingness to disclose AI use, whereas those within the bottom quartile were classified as having low willingness. Interviews were conducted iteratively until thematic saturation was reached, indicating that no substantially new themes emerged from additional interviews. In total, 22 students participated in the semi-structured interviews. The demographic characteristics of the interview participants are presented in Appendix B.

This study followed established ethical guidelines for research involving human participants. Participation in both the questionnaire and interview phases was voluntary, and informed consent was obtained from all participants prior to data collection. Participants were informed about the purpose of the study, their right to withdraw at any time without penalty, and the confidentiality of their responses. Questionnaire responses were collected anonymously, and interview data were anonymised using participant codes to protect participants' identities.

### 5.3 Questionnaires

Data were collected using a structured questionnaire designed to measure the constructs in the proposed conceptual model. The questionnaire included nine constructs: perceived fairness (Chambers et al., 2023), perceived teacher support (Chiu et al., 2023), perceived organisational support (Aldabbas et al., 2025), perceived stigma (Malik et al., 2022), perceived uncertainty (Usman et al., 2021), privacy concern (Zhou & Zhang, 2025), psychological safety (McGuire, 2025), evaluation apprehension (Ismail & Heydarnejad, 2023), and willingness to disclose AI use (Zhou & Wu, 2025b). All measurement items were adapted from established scales in prior research and modified to fit the context of AI use in higher education (Y. Du, 2024). The questionnaire was administered in Chinese to ensure participants' comprehension. To maintain semantic equivalence between the English and Chinese versions, a translation and back-translation procedure was conducted following established guidelines (Brislin, 1970). Prior to the main data collection, a pilot study with 30 university students was conducted to assess the clarity and appropriateness of the questionnaire items, and minor wording revisions were made based on their feedback. In addition to the construct measures, the questionnaire collected participants' demographic information, including gender, age, study level, and academic discipline, and invited respondents to indicate their willingness to participate in a follow-up interview. All items were measured using a five-point Likert scale ranging from 1 (strongly disagree) to 5 (strongly agree). The full list of constructs and measurement items is presented in Appendix C.

### 5.4 Semi-Structured Interviews

To complement the quantitative findings, semi-structured interviews were conducted to obtain a more in-depth understanding of students' perceptions and experiences related to AI use and disclosure in academic contexts. The interview protocol was developed in alignment with the hypotheses of the proposed conceptual model, and the questions were designed to explore participants' views on factors such as perceived fairness, teacher support, organisational support, perceived stigma, perceived uncertainty, privacy concern, psychological safety, and evaluation apprehension in relation to their willingness to disclose AI use. Semi-structured interviews were selected because they allow participants to elaborate on their experiences while ensuring that key issues relevant to the research questions are consistently covered. Participants who had indicated their willingness to take part in a follow-up interview in the questionnaire were subsequently contacted and invited to participate. All interviews were conducted individually in Chinese and lasted approximately 30–45 minutes. After obtaining informed consent, the interviews followed the interview protocol, with additional probing questions used when necessary to clarify responses and encourage further elaboration. The complete interview protocol is provided in Appendix D.

### 5.5 Data Analysis

The quantitative data were analysed using R within a structural equation modelling (SEM) framework (Whittaker & Schumacker, 2022). First, descriptive statistics, including means, standard deviations, skewness, and kurtosis, were calculated to examine the distribution of the variables and assess normality. Confirmatory factor analysis (CFA) was then conducted to evaluate the measurement model, including model fit, indicator reliability, internal consistency reliability, convergent validity, and discriminant validity. Reliability was assessed using Cronbach's α and composite reliability (CR), while convergent validity was evaluated through standardised factor loadings and average variance extracted (AVE). Discriminant validity was examined using the Fornell–Larcker criterion. After establishing satisfactory measurement properties, the structural model was estimated to test the hypothesised relationships among the constructs in the proposed conceptual model. Path coefficients and their significance levels were used to determine whether the proposed hypotheses were supported.

The qualitative interview data were analysed using thematic analysis following a deductive–inductive approach (King et al., 2019). All interviews were conducted and transcribed in Chinese to preserve participants' original meanings. An initial coding framework was developed deductively based on the constructs in the proposed conceptual model. The transcripts were then coded using this preliminary framework while allowing additional themes to emerge inductively from the data. To enhance analytical reliability, two researchers independently coded a subset of the transcripts, and inter-coder agreement was assessed using Cohen's kappa, which indicated substantial agreement ($\kappa = 0.87$) (McHugh, 2012). Discrepancies were resolved through discussion before the refined coding scheme was applied to the remaining transcripts. The resulting codes were subsequently organised into broader themes reflecting students' cognitive evaluations, affective responses, and behavioural intentions related to AI use and disclosure. As the interviews were conducted in Chinese, excerpts presented in the manuscript were translated into English during the reporting stage. The translations were reviewed and verified by two bilingual researchers to ensure accuracy and semantic equivalence.

# 6. Results
## 6.1 Quantitative Results

Descriptive statistics for the study variables are presented in Appendix E. Overall, the means of the constructs ranged from 2.96 to 3.62 on the five-point scale, indicating moderate levels of the measured perceptions among participants. Constructs related to potential concerns about AI use, such as perceived uncertainty, privacy concern, and evaluation apprehension, showed relatively higher mean values, whereas psychological safety and willingness to disclose AI use were comparatively lower. Skewness values ranged from −0.12 to 0.33 and kurtosis values ranged from −0.52 to −0.28, both within the commonly accepted threshold of ±1, indicating that the assumption of normality was satisfied for subsequent structural equation modelling analyses.

Confirmatory factor analysis (CFA) was conducted to evaluate the measurement model. The model demonstrated satisfactory fit to the data, with all fit indices meeting recommended thresholds (see Appendix F). All standardised factor loadings exceeded 0.70 and were statistically significant, indicating adequate indicator reliability. As shown in Appendix G, Cronbach's α and composite reliability (CR) values for all constructs were above the recommended threshold of 0.70, demonstrating good internal consistency. Convergent validity was supported, with average variance extracted (AVE) values exceeding 0.50 for all constructs. Discriminant validity was also established, as the square roots of AVE were greater than the inter-construct correlations in accordance with the Fornell–Larcker criterion (see Appendix H).

The structural model was then estimated to examine the hypothesised relationships among the constructs. As reported in Appendix I and Figure 2, psychological safety had a significant positive effect on willingness to disclose AI use ($\beta = 0.48$, $p < .001$), supporting H1. Perceived fairness ($\beta = 0.29$, $p < .001$), perceived teacher support ($\beta = 0.34$, $p < .001$), and perceived organisational support ($\beta = 0.21$, $p < .01$) were all positively associated with psychological safety, supporting H2–H4. In addition, evaluation apprehension negatively predicted willingness to disclose AI use ($\beta = -0.31$, $p < .001$) and psychological safety ($\beta = -0.26$, $p < .001$), supporting H5 and H6. Finally, perceived stigma ($\beta = 0.41$, $p < .001$), perceived uncertainty ($\beta = 0.28$, $p < .001$), and privacy concern ($\beta = 0.24$, $p < .01$) were positively associated with evaluation apprehension, supporting H7–H9. Overall, the quantitative results provided empirical support for the proposed conceptual model.

**Figure 2. Structural Model Results**

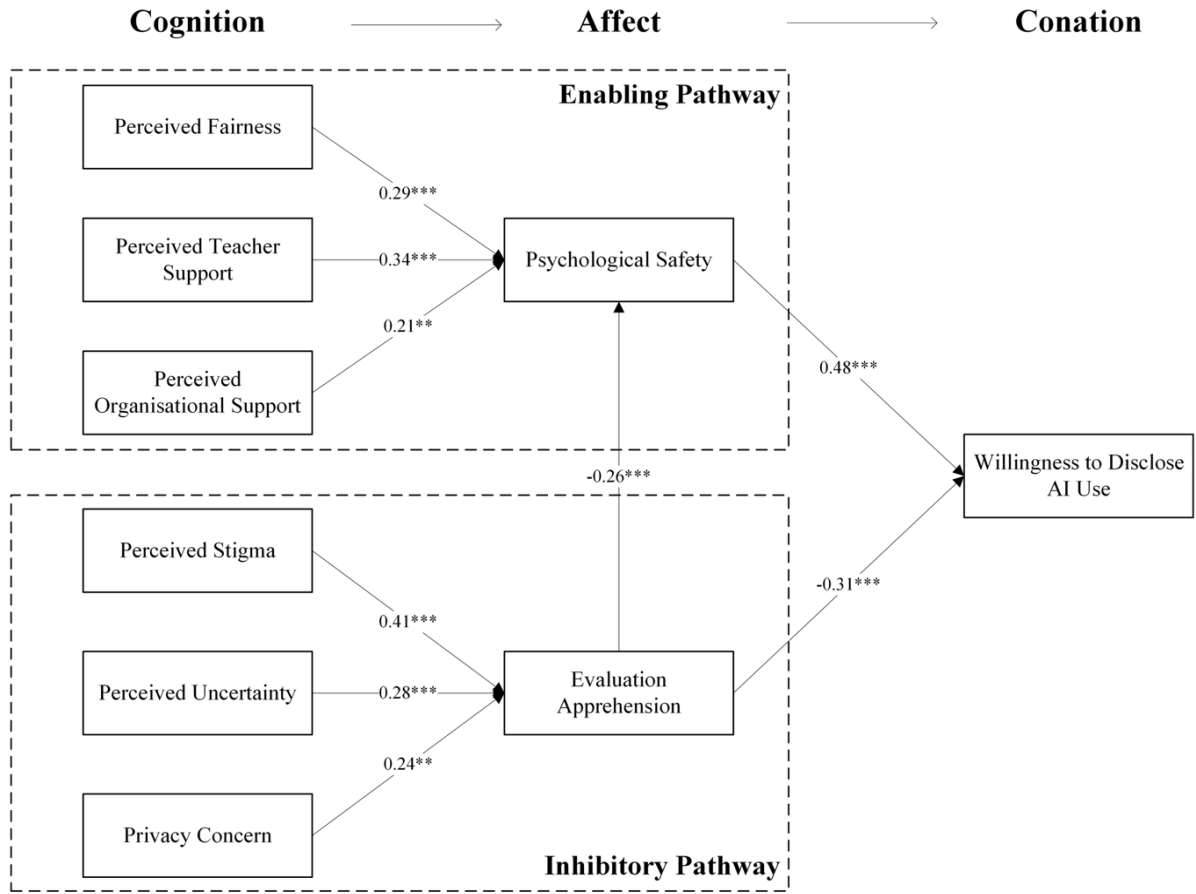

Note. Statistical significance is denoted as *** $p < .001$, ** $p < .01$.

## 6.2 Qualitative Results

The qualitative interviews were conducted to deepen understanding of the factors influencing students' willingness to disclose AI use and to contextualise the quantitative findings. Thematic analysis of the interview data identified three major themes: (1) institutional clarity and support fostering openness, (2) stigma and uncertainty generating concerns about evaluation, and (3) conditional and strategic disclosure practices. These themes illustrate how students interpret their academic environment and how such interpretations shape their decisions about whether to disclose the use of AI tools.

The first theme concerns the role of institutional clarity and instructional support in encouraging openness about AI use. Many participants reported that they would feel more comfortable disclosing AI use when institutional policies and course expectations were clearly communicated. In such situations, students tended to perceive AI use as a recognised learning tool rather than a behaviour associated with academic misconduct. For example, P07 explained that disclosure becomes easier "if the teacher clearly explains how AI can be used, such as for brainstorming or checking language." Similarly, P03 noted that when instructors explicitly address AI use in assignment guidelines, "students feel safer mentioning it because the rules are already clear." However, participants also pointed out that such clarity was not always consistent across courses. Several students described receiving mixed messages from different instructors regarding the legitimacy of AI use. As P02 commented, "some teachers encourage using AI as a learning tool, while others strongly discourage it." This inconsistency often created hesitation, even among students who were otherwise willing to be transparent.

The second theme relates to concerns about negative evaluation arising from stigma and policy ambiguity. Many students perceived that AI use might still be associated with academic dishonesty or reduced academic ability. As a result, disclosing AI use could potentially expose them to negative judgement from instructors. P15 described this concern by explaining that "even if AI only helps with ideas, teachers might assume the whole assignment was generated by AI." Participants also highlighted that institutional policies regarding AI use are still evolving, which makes it difficult to determine when disclosure is expected. P09 observed that "the rules about AI are not very stable yet, so it feels risky to talk about it openly." In addition, a number of students mentioned privacy-related concerns. For instance, P18 suggested that disclosing AI use might lead to closer scrutiny of their work: "if you say you used AI, the teacher might check your assignment more carefully." These concerns often led students to avoid disclosure in order to minimise potential academic risks.

The third theme highlights strategic and conditional approaches to disclosure. Rather than treating disclosure as a simple yes-or-no decision, many participants described making context-dependent judgements about when disclosure would be appropriate. Students frequently differentiated between minor and substantial forms of AI assistance. For example, P20 indicated that they would disclose AI use "if it was only used to check grammar or organise ideas," but would hesitate to disclose when AI contributed more extensively to writing. Similarly, P14 noted that disclosure decisions often depend on the instructor's perceived attitude toward AI: "if the teacher is supportive, I might mention it, but otherwise I would rather not take the risk." Such accounts suggest that students often evaluate the potential consequences of disclosure before deciding how transparent to be. Even participants who generally supported transparency acknowledged that disclosure could be risky in high-stakes assessment contexts. As P10 explained, "being honest about AI use sounds reasonable, but you still worry about how the teacher might interpret it."

Overall, the qualitative findings provide additional insight into the complex considerations underlying students' disclosure decisions. While supportive policies and clear communication from instructors can encourage openness about AI use, persistent stigma, policy ambiguity, and concerns about evaluation may lead students to adopt cautious or selective disclosure strategies. These findings help explain why students' willingness to disclose AI use varies across contexts and academic environments.

## 7. Discussion
### 7.1 Enabling Pathway of Willingness to Disclose AI Use
The quantitative findings provide clear support for the enabling pathway proposed in the Cognition–Affect–Conation framework, demonstrating that psychological safety is a central mechanism shaping students' willingness to disclose AI use. Structural modelling showed that psychological safety significantly predicted disclosure intention, while perceived fairness, teacher support, and organisational support all positively influenced psychological safety. These results indicate that students' cognitive evaluations of institutional practices translate into affective states that shape behavioural intentions, consistent with CAC theory (Zhou & Zhang, 2024). When students perceive policies and assessment practices as fair, they are less likely to anticipate punitive or biased evaluation, which strengthens their sense of interpersonal security and openness (Chambers et al., 2023; Malhotra et al., 2022). Similarly, supportive instructor behaviour and institutional guidance appear to function as relational signals that legitimise AI-assisted learning, thereby fostering an environment in which disclosure becomes psychologically safer. This finding aligns with prior work showing that supportive educational climates increase students' willingness to communicate learning practices and engage in transparent academic behaviour (Held & Mori, 2024; McGuire, 2025). Nevertheless, the moderate magnitude of these effects suggests that institutional support alone may not fully normalise AI disclosure, indicating that students' perceptions of the broader academic culture remain an important contextual factor.

The qualitative findings further illuminate how these enabling conditions operate in practice while also revealing important limitations. Interview participants consistently described institutional clarity and supportive instructor attitudes as factors that increase their comfort in acknowledging AI use, reinforcing the quantitative evidence that fairness and support enhance psychological safety. When instructors explicitly explain acceptable AI practices or integrate AI guidance into assignment instructions, students interpret such signals as institutional endorsement of responsible AI use, reducing perceived risk associated with disclosure. This observation is consistent with research suggesting that clear institutional guidance and ethical framing can normalise emerging technologies in educational settings (Francis et al., 2025; Jensen et al., 2025). However, the interviews also reveal that these enabling mechanisms are often undermined by inconsistency across courses and instructors. Students reported receiving conflicting messages about whether AI use is acceptable, which weakens the stabilising effect of perceived fairness and support. This suggests that psychological safety may be highly context-dependent, emerging not only from supportive actors but also from the coherence of institutional policies. Consequently, while the enabling pathway identified in the quantitative model is empirically supported, the qualitative evidence indicates that its effectiveness depends on consistent institutional communication and pedagogical alignment regarding AI use.

### 7.2 Inhibitory Pathway of Willingness to Disclose AI Use

The quantitative results also provide strong support for the inhibitory pathway proposed in the CAC framework. Evaluation apprehension significantly reduced students' willingness to disclose AI use and simultaneously undermined psychological safety, indicating that concerns about negative judgement function as a critical emotional barrier to transparency. This finding is consistent with prior research suggesting that evaluation apprehension discourages individuals from revealing behaviours that may expose them to criticism or reputational risk (Ismail & Heydarnejad, 2023; McGaghie, 2018). In the present study, cognitive perceptions of stigma, uncertainty, and privacy concern significantly increased evaluation apprehension, suggesting that students' interpretation of the academic environment can activate anticipatory anxiety about how their AI use will be judged. When AI-assisted work is perceived as socially questionable or ethically ambiguous, students may avoid disclosure as a protective strategy to minimise potential academic consequences. This result aligns with emerging literature indicating that ambiguity surrounding AI policies and norms can create psychological tension that discourages open reporting of AI use (Stone, 2025; Yan et al., 2025). Moreover, the negative association between evaluation apprehension and psychological safety highlights the dual role of this affective factor: it not only directly suppresses disclosure intentions but also indirectly weakens the supportive emotional conditions required for transparent behaviour.

The qualitative findings further illuminate how these inhibitory mechanisms manifest in students' decision-making processes. Many interview participants described AI use as carrying implicit stigma within academic contexts, particularly when instructors equate AI assistance with academic dishonesty or reduced intellectual effort. Such perceptions intensified students' concerns that disclosure might lead instructors to question the originality or quality of their work, thereby reinforcing evaluation apprehension. Participants also emphasised the role of policy uncertainty, noting that rapidly evolving institutional guidelines make it difficult to determine when AI use is acceptable or how it should be reported. Similar concerns have been identified in studies showing that ambiguous technological regulations can heighten perceived risk and discourage open communication about emerging practices (Usher & Barak, 2024; Zhu et al., 2025). Privacy-related worries further compounded these concerns, as some students feared that disclosing AI use could trigger closer scrutiny of their assignments or expose their learning processes to institutional monitoring (Carmody et al., 2021; Hu & Min, 2023). Importantly, the interviews suggest that these inhibitory pressures often lead students to adopt cautious or strategic disclosure practices rather than complete non-disclosure, indicating that transparency about AI use is shaped by a dynamic risk–benefit assessment rather than a purely normative decision.

### 7.3 Theoretical and Practical Implications

This study contributes theoretically by extending the Cognition–Affect–Conation (CAC) framework to explain students' willingness to disclose AI use in higher education. While prior research on AI in education has mainly examined adoption or ethical concerns, limited attention has been given to the psychological mechanisms underlying disclosure behaviour (Yan et al., 2025; Zhou & Wu, 2025b). The findings demonstrate that disclosure intentions emerge through a cognition–affect process in which institutional perceptions shape emotional responses that subsequently influence behavioural intentions. In particular, the results highlight the mediating role of psychological safety and evaluation apprehension, reinforcing the explanatory value of the CAC perspective in technology-related educational behaviour (Zhou & Zhang, 2024).

Practically, the findings suggest that universities and instructors play a central role in fostering transparency about AI use. Clear and consistent institutional policies are necessary to reduce ambiguity and strengthen students' perceptions of fairness and security (Francis et al., 2025; Stone, 2025). In addition, supportive instructor communication and explicit guidance on acceptable AI practices may reduce evaluation concerns and normalise disclosure. Integrating AI literacy and ethical guidance into curricula may further help reduce stigma associated with AI-assisted learning and encourage responsible disclosure practices (Wiese et al., 2025).

### 7.4 Limitations and Future Directions
This study has several limitations that should be considered when interpreting the findings. First, the data were collected from university students in China using a convenience sampling strategy, which may limit the generalisability of the results to other educational contexts or cultural settings (Y. Du et al., 2024; Y. Du, 2025; Y. Du, Yuan, et al., 2026). Institutional norms and attitudes toward AI use may vary across countries and universities, potentially influencing students' perceptions and disclosure behaviour. Second, although the study employed a mixed-methods design, the quantitative data were cross-sectional, which restricts the ability to establish causal relationships among the constructs (Y. Du et al., 2025; Zhang et al., 2026). Finally, the measures relied on self-reported perceptions and intentions, which may be influenced by social desirability or respondents' subjective interpretations of AI use in academic tasks (Y. Du & He, 2026a, 2026b; He & Du, 2024).

Future research could address these limitations in several ways. Comparative studies across different countries or institutional contexts would help examine how cultural and policy environments influence students' willingness to disclose AI use (Y. Du, 2026; Y. Du, Li, et al., 2026; Wang et al., 2026). Longitudinal or experimental designs could also provide stronger evidence regarding the causal mechanisms linking cognitive perceptions, affective responses, and disclosure behaviour (C. Du et al., 2025; Wang et al., 2024; Zou et al., 2023, 2024). In addition, future studies may explore how specific pedagogical practices, such as AI-integrated assessment design or explicit disclosure guidelines, shape students' transparency regarding AI-assisted work. Such research could provide deeper insights into how educational institutions can effectively promote responsible and open AI use in academic environments.

### 8. Conclusion
This study examined university students' willingness to disclose AI use through the Cognition–Affect–Conation framework by identifying both enabling and inhibitory psychological pathways. The findings show that psychological safety plays a central role in encouraging disclosure, shaped by students' perceptions of fairness, teacher support, and organisational support. At the same time, evaluation apprehension acts as a key barrier, emerging from perceived stigma, uncertainty, and privacy concerns associated with AI use. Together, the results suggest that students' disclosure decisions are influenced by the interaction between supportive institutional conditions and perceived risks of negative evaluation. By integrating quantitative and qualitative evidence, this study provides a more comprehensive understanding of the psychological mechanisms underlying AI-use transparency in higher education and highlights the importance of clear policies, supportive teaching practices, and reduced stigma in promoting responsible disclosure of AI-assisted work.

**Appendices**

**Appendix A. Questionnaire Participant Characteristics (*N* = 546)**

| Characteristic | Category | *n* | % |
| --- | --- | --- | --- |
| Gender | Male | 252 | 46.2 |
| | Female | 294 | 53.8 |
| Age | 18–20 | 198 | 36.3 |
| | 21–23 | 257 | 47.1 |
| | 24 or above | 91 | 16.7 |
| Study level | Undergraduate | 421 | 77.1 |
| | Postgraduate | 125 | 22.9 |
| Academic discipline | STEM | 274 | 50.2 |
| | Non-STEM | 272 | 49.8 |

**Appendix B. Interview Participant Characteristics (*N* = 22)**

| ID | Gender | Age | Study level | Academic discipline | Willingness to disclose AI use |
| --- | --- | --- | --- | --- | --- |
| P01 | Female | 21 | Undergraduate | Non-STEM | High |
| P02 | Male | 23 | Undergraduate | STEM | Low |
| P03 | Female | 22 | Undergraduate | Non-STEM | High |
| P04 | Male | 24 | Postgraduate | STEM | Low |
| P05 | Female | 20 | Undergraduate | Non-STEM | High |
| P06 | Male | 25 | Postgraduate | STEM | Low |
| P07 | Female | 21 | Undergraduate | STEM | High |
| P08 | Male | 22 | Undergraduate | Non-STEM | Low |
| P09 | Female | 23 | Undergraduate | STEM | High |
| P10 | Male | 26 | Postgraduate | STEM | Low |
| P11 | Female | 22 | Undergraduate | Non-STEM | High |
| P12 | Male | 24 | Postgraduate | STEM | Low |
| P13 | Female | 21 | Undergraduate | Non-STEM | High |
| P14 | Male | 23 | Undergraduate | STEM | Low |
| P15 | Female | 22 | Undergraduate | Non-STEM | High |
| P16 | Male | 25 | Postgraduate | STEM | Low |
| P17 | Female | 20 | Undergraduate | Non-STEM | High |
| P18 | Male | 24 | Postgraduate | STEM | Low |
| P19 | Female | 23 | Undergraduate | STEM | High |
| P20 | Male | 22 | Undergraduate | Non-STEM | Low |
| P21 | Female | 21 | Undergraduate | Non-STEM | High |
| P22 | Male | 26 | Postgraduate | STEM | Low |

**Appendix C. Constructs and Measurement Items**

| Construct | Item | Measurement Item (English) | Measurement Item (Chinese) |
| --- | --- | --- | --- |
| Perceived Fairness (Chambers et al., 2023) | PF1 | The rules regarding AI use in academic work at my university are fair. | 我认为我所在大学关于学术作业中使用人工智能的规定是公平的。 |
| | PF2 | The policies about AI use are applied consistently to students. | 关于人工智能使用的政策在学生之间是一致执行的。 |
| | PF3 | The expectations about AI use in coursework are reasonable. | 课程中关于人工智能使用的要求是合理的。 |
| Perceived Teacher Support (Chiu et al., 2023) | PTS1 | My instructors provide guidance on appropriate AI use in academic work. | 我的教师会指导我们如何在学术作业中合理使用人工智能。 |
| | PTS2 | My instructors are open to discussing the use of AI tools in learning. | 我的教师愿意与学生讨论在学习中使用人工智能工具。 |
| | PTS3 | My instructors support students in learning to use AI responsibly. | 我的教师支持学生学习负责任地使用人工智能。 |
| Perceived Organisational Support (Aldabbas et al., 2025) | POS1 | My university provides clear guidance on AI use in academic work. | 我所在大学为学术作业中的人工智能使用提供了清晰的指导。 |
| | POS2 | My university provides resources to help students understand AI use in learning. | 我所在大学提供资源帮助学生了解如何在学习中使用人工智能。 |
| | POS3 | My university supports students in adapting to AI-related academic practices. | 我所在大学支持学生适应与人工智能相关的学习方式。 |
| Psychological Safety (McGuire, 2025) | PS1 | I feel safe acknowledging my use of AI tools in academic work. | 在学术作业中承认使用人工智能工具时，我感到是安全的。 |
| | PS2 | I can openly discuss my use of AI tools in coursework. | 我可以公开讨论自己在课程作业中使用人工智能工具的情况。 |
| | PS3 | I feel comfortable being transparent about my use of AI in assignments. | 对于在作业中说明自己使用人工智能，我感到很自在。 |
| Perceived Stigma (Malik et al., 2022) | PST1 | Using AI tools in academic work may lead others to view me negatively. | 在学术作业中使用人工智能可能会让他人对我产生负面看法。 |
| | PST2 | Students who use AI tools may be judged negatively by instructors or peers. | 使用人工智能工具的学生可能会受到教师或同学的负面评价。 |
| | PST3 | Using AI tools in assignments may harm a student's academic reputation. | 在作业中使用人工智能可能会影响学生的学术声誉。 |
| Perceived Uncertainty (Usman et al., 2021) | PU1 | I am unsure when AI tools are allowed in academic work. | 我不确定在什么情况下可以在学术作业中使用人工智能工具。 |
| | PU2 | The rules about AI use in coursework are unclear to me. | 我觉得课程中关于人工智能使用的规定并不清晰。 |
| | PU3 | I am uncertain about the consequences of using AI tools in assignments. | 我不确定在作业中使用人工智能工具可能带来的后果。 |
| Privacy Concern (Zhou & Zhang, 2025) | PC1 | I worry that disclosing my AI use may expose my learning processes. | 我担心披露自己使用人工智能会暴露我的学习过程。 |
| | PC2 | I am concerned about how information about my AI use might be used by instructors or institutions. | 我担心教师或学校可能会如何使用我披露的人工智能使用信息。 |
| | PC3 | I am concerned about the privacy of my academic work when AI use is disclosed. | 当需要披露人工智能使用情况时，我会担心学术作业的隐私问题。 |
| Evaluation Apprehension (Ismail & Heydarnejad, 2023) | EA1 | I worry that instructors may judge me negatively if I disclose my AI use. | 如果我披露使用人工智能工具，我会担心教师对我产生负面评价。 |
| | EA2 | I feel anxious about how others might evaluate my AI use in academic work. | 我会担心他人如何评价我在学术作业中使用人工智能。 |
| | EA3 | I feel uneasy about disclosing my use of AI tools. | 对于披露自己使用人工智能工具，我会感到不安。 |
| Willingness to Disclose AI Use (Zhou & Wu, 2025b) | WD1 | I would report my use of AI tools when completing academic assignments. | 在完成学术作业时，我愿意报告自己使用了人工智能工具。 |
| | WD2 | I would acknowledge AI assistance when submitting coursework. | 在提交课程作业时，我愿意说明人工智能的辅助作用。 |
| | WD3 | I would disclose my use of AI tools to my instructors when required. | 在需要时，我愿意向教师披露自己使用人工智能工具的情况。 |

**Appendix D. Semi-Structured Interview Protocol**

| Hypothesis | Interview Question (English) | Interview Question (Chinese) |
|---|---|---|
| H1 Psychological Safety → Willingness to Disclose | How comfortable do you feel acknowledging your use of AI tools in your academic work? What factors influence whether you disclose this use? | 在学术作业中承认自己使用人工智能工具时，你感觉有多自在？哪些因素会影响你是否愿意披露这种使用？ |
| H2 Perceived Fairness → Psychological Safety | How do you perceive the fairness of university policies regarding AI use in academic work? How do these policies affect your willingness to be open about using AI? | 你如何看待学校关于学术作业中使用人工智能的相关政策是否公平？这些政策会如何影响你是否愿意公开自己使用人工智能？ |
| H3 Perceived Teacher Support → Psychological Safety | In what ways do your instructors influence how comfortable you feel discussing or disclosing AI use in coursework? | 教师在多大程度上会影响你讨论或披露自己使用人工智能工具的意愿？ |
| H4 Perceived Organisational Support → Psychological Safety | How does your university support or guide students regarding AI use in learning? How does this affect your willingness to disclose AI use? | 学校在学习中如何为学生提供关于人工智能使用的指导或支持？这些支持是否会影响你披露人工智能使用情况的意愿？ |
| H5 Evaluation Apprehension → Willingness to Disclose | Do you worry about being negatively evaluated if you disclose your use of AI tools? Can you describe situations where such concerns might arise? | 如果你披露自己使用人工智能工具，你是否担心受到负面评价？能否举例说明这种担忧可能在什么情况下出现？ |
| H6 Evaluation Apprehension → Psychological Safety | How do concerns about evaluation from instructors or peers affect how safe you feel disclosing your AI use? | 来自教师或同学评价的担忧会如何影响你披露人工智能使用时的安全感？ |
| H7 Perceived Stigma → Evaluation Apprehension | In your view, how do students and instructors generally perceive the use of AI in academic work? Do you think using AI carries any stigma? | 在你看来，学生和教师通常如何看待在学术作业中使用人工智能？你认为这种行为会带来某种"污名化"吗？ |

| | | | | |
|---|---|---|---|---|
| H8 Perceived Uncertainty → Evaluation Apprehension | Are the rules about AI use in academic work clear to you? How does uncertainty about these rules influence your decisions about disclosing AI use? | | 你觉得关于学术作业中使用人工智能的规定是否清晰？这种不确定性会如何影响你是否披露人工智能使用？ | |
| H9 Privacy Concern → Evaluation Apprehension | Do you have any concerns about privacy or personal information when disclosing your use of AI tools? | | 当披露自己使用人工智能工具时，你是否会担心隐私或个人信息问题？ | |
| Overall Disclosure Behaviour | Can you describe a situation where you used AI in academic work and decided whether or not to disclose it? What influenced your decision? | | 你能描述一次在学术作业中使用人工智能并决定是否披露的经历吗？哪些因素影响了你的决定？ | |

Note. The questions served as guiding prompts during the interviews. Follow-up questions were asked when necessary to encourage participants to elaborate on their experiences and perspectives.

**Appendix E. Descriptive Statistics of the Constructs**

| Construct | M | SD | Skewness | Kurtosis |
|---|---|---|---|---|
| Perceived Fairness | 3.28 | 0.84 | -0.12 | -0.41 |
| Perceived Teacher Support | 3.41 | 0.80 | -0.09 | -0.46 |
| Perceived Organisational Support | 3.19 | 0.83 | -0.05 | -0.52 |
| Perceived Stigma | 3.46 | 0.87 | 0.18 | -0.37 |
| Perceived Uncertainty | 3.58 | 0.90 | 0.24 | -0.43 |
| Privacy Concern | 3.62 | 0.85 | 0.29 | -0.39 |
| Psychological Safety | 3.07 | 0.82 | 0.11 | -0.49 |
| Evaluation Apprehension | 3.54 | 0.88 | 0.26 | -0.34 |
| Willingness to Disclose AI Use | 2.96 | 0.91 | 0.33 | -0.28 |

**Appendix F. Model Fit Indices**

| Fit Index | Threshold | Measurement Model | Structural Model |
|---|---|---|---|
| $\chi^2/df$ | < 3.00 | 2.31 | 2.45 |
| CFI | ≥ 0.90 | 0.94 | 0.93 |
| TLI | ≥ 0.90 | 0.93 | 0.92 |
| RMSEA | ≤ 0.08 | 0.049 | 0.052 |
| SRMR | ≤ 0.08 | 0.046 | 0.051 |

**Appendix G. Reliability and Convergent Validity**

| Construct | Item | Factor Loading | Cronbach's α | CR | AVE |
|---|---|---|---|---|---|
| Perceived Fairness | PF1 | 0.81 | 0.86 | 0.88 | 0.71 |
| | PF2 | 0.85 | | | |
| | PF3 | 0.84 | | | |
| Perceived Teacher Support | PTS1 | 0.83 | 0.88 | 0.90 | 0.74 |
| | PTS2 | 0.87 | | | |
| | PTS3 | 0.86 | | | |
| Perceived Organisational Support | POS1 | 0.79 | 0.85 | 0.87 | 0.69 |
| | POS2 | 0.84 | | | |
| | POS3 | 0.83 | | | |
| Perceived Stigma | PST1 | 0.82 | 0.87 | 0.89 | 0.73 |
| | PST2 | 0.88 | | | |
| | PST3 | 0.84 | | | |
| Perceived Uncertainty | PU1 | 0.80 | 0.86 | 0.88 | 0.70 |
| | PU2 | 0.86 | | | |
| | PU3 | 0.85 | | | |
| Privacy Concern | PC1 | 0.81 | 0.85 | 0.87 | 0.69 |
|  | PC2 | 0.85 |  |  |  |
|  | PC3 | 0.83 |  |  |  |
| Psychological Safety | PS1 | 0.84 | 0.89 | 0.91 | 0.77 |
|  | PS2 | 0.88 |  |  |  |
|  | PS3 | 0.89 |  |  |  |
| Evaluation Apprehension | EA1 | 0.83 | 0.87 | 0.89 | 0.73 |
|  | EA2 | 0.87 |  |  |  |
|  | EA3 | 0.85 |  |  |  |
| Willingness to Disclose AI Use | WD1 | 0.86 | 0.90 | 0.92 | 0.79 |
|  | WD2 | 0.89 |  |  |  |
|  | WD3 | 0.90 |  |  |  |

**Appendix H. Discriminant Validity (Fornell–Larcker Criterion)**

| Construct | PF | PTS | POS | PST | PU | PC | PS | EA | WD |
|---|---|---|---|---|---|---|---|---|---|
| PF | **0.84** |  |  |  |  |  |  |  |  |
| PTS | 0.54 | **0.86** |  |  |  |  |  |  |  |
| POS | 0.49 | 0.57 | **0.83** |  |  |  |  |  |  |
| PST | -0.31 | -0.28 | -0.26 | **0.85** |  |  |  |  |  |
| PU | -0.22 | -0.19 | -0.21 | 0.46 | **0.84** |  |  |  |  |
| PC | -0.18 | -0.17 | -0.20 | 0.39 | 0.44 | **0.83** |  |  |  |
| PS | 0.58 | 0.61 | 0.55 | -0.36 | -0.29 | -0.24 | **0.88** |  |  |
| EA | -0.35 | -0.32 | -0.30 | 0.52 | 0.47 | 0.43 | -0.41 | **0.85** |  |
| WD | 0.49 | 0.45 | 0.42 | -0.38 | -0.33 | -0.29 | 0.57 | -0.46 | **0.89** |

Note. Diagonal elements (in bold) represent the square root of the average variance extracted (AVE). Off-diagonal elements represent correlations among constructs. PF = perceived fairness; PTS = perceived teacher support; POS = perceived organisational support; PST = perceived stigma; PU = perceived uncertainty; PC = privacy concern; PS = psychological safety; EA = evaluation apprehension; WD = willingness to disclose AI use.

**Appendix I. Structural Model Results**

| Hypothesis | Path | $\beta$ | SE | z | Result |
|---|---|---|---|---|---|
| H1 | PS → WD | 0.48 | 0.05 | 9.60*** | Supported |
| H2 | PF → PS | 0.29 | 0.06 | 4.83*** | Supported |
| H3 | PTS → PS | 0.34 | 0.05 | 6.80*** | Supported |
| H4 | POS → PS | 0.21 | 0.06 | 3.50** | Supported |
| H5 | EA → WD | -0.31 | 0.06 | -5.17*** | Supported |
| H6 | EA → PS | -0.26 | 0.05 | -5.20*** | Supported |
| H7 | PST → EA | 0.41 | 0.05 | 8.20*** | Supported |
| H8 | PU → EA | 0.28 | 0.06 | 4.67*** | Supported |
| H9 | PC → EA | 0.24 | 0.07 | 3.43** | Supported |

Note. PF = perceived fairness; PTS = perceived teacher support; POS = perceived organisational support; PST = perceived stigma; PU = perceived uncertainty; PC = privacy concern; PS = psychological safety; EA = evaluation apprehension; WD = willingness to disclose AI use. Statistical significance is denoted as *** $p < .001$, ** $p < .01$.